\documentclass[10pt,twocolumn,letterpaper]{article}

\usepackage[pagenumbers]{cvpr} 

\usepackage{amsmath,amssymb,amsthm}
\theoremstyle{definition}

\usepackage{booktabs}
\usepackage{adjustbox}
\usepackage{array}
\usepackage{multirow}

\newcommand{\ignore}[1]{}

\usepackage[most]{tcolorbox}  

\tcbset{
    myhypothesis/.style={
        colback=gray!10,       
        colframe=gray!50,      
        boxrule=0.8pt,         
        arc=4pt,               
        left=4pt, right=4pt, top=4pt, bottom=4pt, 
        fonttitle=\bfseries,
        coltitle=black,
        enhanced,
    }
}

\definecolor{cvprblue}{rgb}{0.21,0.49,0.74}
\usepackage[pagebackref,breaklinks,colorlinks,allcolors=cvprblue]{hyperref}

\def\paperID{16639} 
\def\confName{CVPR}
\def\confYear{2026}

\title{The Universal Normal Embedding}

\author{
\makebox[\textwidth][c]{%
\begin{tabular}{c}
Chen Tasker\thanks{These authors contributed equally to this work.\\ Corresponding author: \texttt{roybe@campus.technion.ac.il}}
\quad Roy Betser\footnotemark[1] \quad
Eyal Gofer\footnotemark[1] \quad
Meir Yossef Levi \quad
Guy Gilboa \\
Viterbi Faculty of Electrical and Computer Engineering\\
Technion - Israel Institute of Technology
\end{tabular}%
}
}

\begin{document}
\maketitle


\begin{abstract}
Generative models and vision encoders have largely advanced on separate tracks, optimized for different goals and grounded in different mathematical principles. 
Yet, they share a fundamental property: latent space Gaussianity. 
Generative models map Gaussian noise to images, while encoders map images to semantic embeddings whose coordinates empirically behave as Gaussian.
We hypothesize that both are views of a shared latent source, the \emph{Universal Normal Embedding (UNE)}: an approximately Gaussian latent space from which encoder embeddings and DDIM-inverted noise arise as noisy linear projections.
To test our hypothesis, we introduce \emph{NoiseZoo}, a dataset of per-image latents comprising DDIM-inverted diffusion noise and matching encoder representations (CLIP, DINO).
On CelebA, linear probes in both spaces yield strong, aligned attribute predictions, indicating that generative noise encodes meaningful semantics along linear directions.
These directions further enable faithful, controllable edits (e.g., smile, gender, age) without architectural changes, where simple orthogonalization mitigates spurious entanglements.
Taken together, our results provide empirical support for the UNE hypothesis and reveal a shared Gaussian-like latent geometry that concretely links encoding and generation. Code and data are available \href{https://rbetser.github.io/UNE/}{here}. 
\end{abstract}

\section{Introduction}
Generative modeling has reshaped visual computing, enabling high-fidelity synthesis, reconstruction, and editing~\citep{goodfellow2014generative,kingma2013auto,ho2020denoising,rombach2022high}. 
In parallel, foundation models have learned highly semantic representations through self-supervision, where simple linear heads achieve strong classification, retrieval, and zero-shot recognition~\citep{chen2020simple,caron2021emerging,radford2021learning}. 
Together, these advances shifted vision from passive recognition to general-purpose creation and understanding, now spanning diverse visual domains~\citep{patashnik2021styleclip,avrahami2022blended}.

Prior work reveals surprising \emph{linearity} and even shared geometry across deep latent spaces~\citep{zhang2024emergence, badrinath2024all}. 
First, within \emph{generative} families, independently trained VAEs, GANs, flows, and diffusion models can be ``stitched'' (i.e., their latent spaces can be linearly aligned so that codes from one model can be decoded by another) via simple linear maps between their latents~\citep{asperti2023comparing,badrinath2024all,lahner2023direct,maiorca2023latent,zhang2024emergence}.
Similarly, within \emph{representation} families, vision encoders likewise ``stitch’’ across architectures and modalities.
Single-projection text-image alignment and shallow model-stitching show that independently trained encoders can operate in a shared latent space~\citep{merullo2022linearly,bansal2021revisiting,liu2023visual,wang2024qwen2,kornblith2019similarity}.

\begin{figure}[t]
    \centering
    \includegraphics[width=0.75\linewidth]{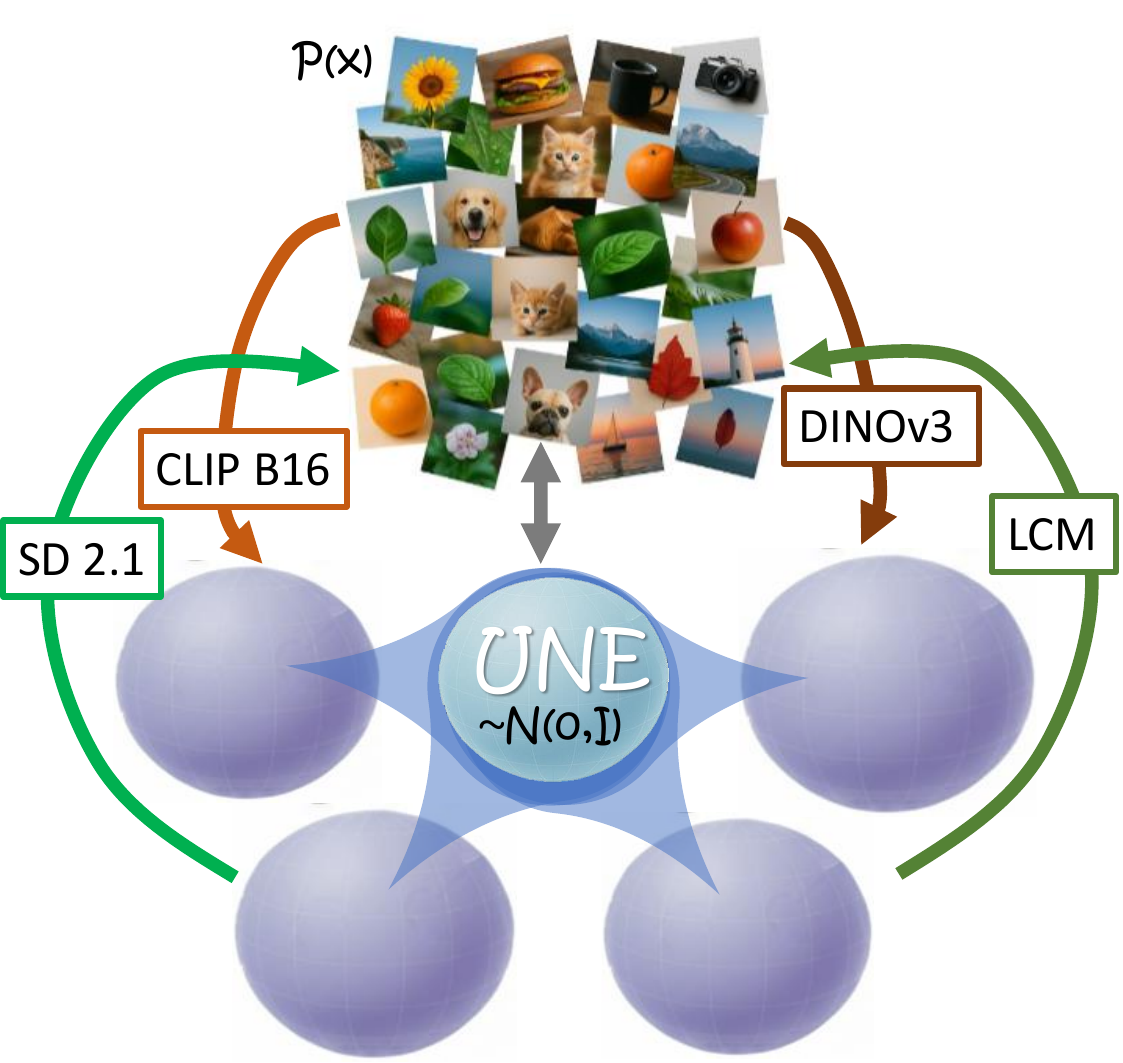}
    \caption{\textbf{UNE conceptual illustration.}
    Different encoders (e.g., CLIP, DINO) and generative models (e.g., SD, LCM) provide different views of the same underlying Gaussian latent structure.
    Although trained for different objectives, their latents can be interpreted as noisy linear projections of a shared ideal Gaussian space.}
    \label{fig:teaser}
\end{figure}

\begin{figure*}[t]
    \centering
    \includegraphics[width=0.95\linewidth]{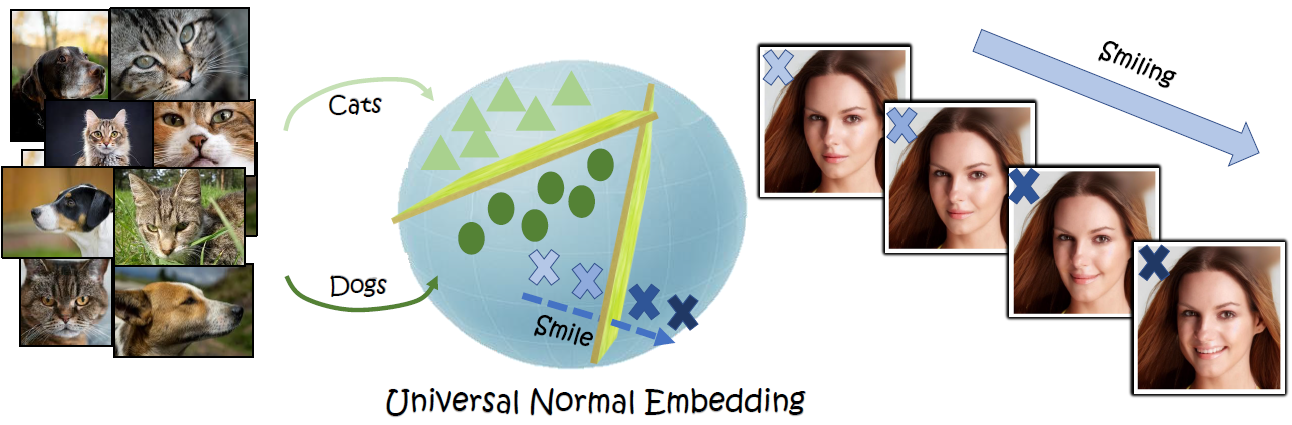}      
    \caption{\textbf{Universal Normal Embedding (UNE).}
    A multivariate standard Gaussian latent space representing the encoded data distribution, in which linear directions align with semantics: classes are separable by hyperplanes, and continuous attributes (e.g., ``smile'') can be edited by perturbing along a single latent direction.}
    \label{fig:une}
\end{figure*}


Motivated by the Platonic Representation Hypothesis and embedding-translation results~\citep{huh2024platonic,jha2025harnessing}, and by identifiability showing that contrastive encoders invert the data-generating process~\citep{zimmermann2021contrastive}, we unify the encoder and generator worlds by directly linking generative noise to encoder representations.
We posit a shared, approximately Gaussian latent space, the \emph{Universal Normal Embedding (UNE)}, from which both families arise as \emph{noisy linear projections}.
UNE refers to an ideal Gaussian latent space whose linear projections approximate the latent spaces of both generative models and vision encoders (see illustration in \Cref{fig:teaser}).
In this geometry, semantic variation aligns with linear directions~\citep{bishop2006pattern}, making UNE \emph{actionable} for linear probes and controllable edits (illustrated in \Cref{fig:une}).
 
Evidence motivating UNE comes from both sides.
Generative models sample from Gaussian priors, while encoder representations (e.g., CLIP~\citep{radford2021learning}, DINO~\citep{caron2021emerging}) empirically behave as approximately Gaussian~\citep{betser2026general, hayun2026trainingfree}. 
Contrastive-learning theory shows that encoders can recover the latent generative factors~\citep{zimmermann2021contrastive}, and follow-up work establishes identifiability of encoder representations up to linear transformations~\citep{daunhawer2023identifiability,reizinger2024cross}.
In parallel, large models converge toward shared latent geometry across architectures and modalities~\citep{huh2024platonic,jha2025harnessing,tyshchuk2023isotropy,li2020sentence}.
Recent theoretical work further formalizes different regimes in which representations exhibit Gaussian behavior~\citep{balestriero2025lejepa, betser2026infonce}.
These results suggest that encoder latents and generative noise reflect the same underlying factors.
We show that these factors admit an approximately Gaussian shared latent space in practice, with encoders and generators aligning as noisy linear projections of that space.

Having established the motivation and formulation of UNE, we investigate it empirically by analyzing latent representations from multiple diffusion models and vision encoders using a unified per-image dataset.
We evaluate observable consequences predicted by the hypothesis: Gaussianity of coordinates, linear separability of semantic attributes, cross-model latent alignment, and linear controllability of semantic directions.
We further examine multi-view intersections of these latent spaces to study whether they preserve a consistent shared structure.
Together, these evaluations suggest that encoder and generative latents behave as noisy linear views of a common, approximately Gaussian latent source.

\noindent Our main contributions are:
\begin{itemize}
\item \textbf{Universal Normal Embedding (UNE).}
We formalize the UNE hypothesis of a shared, approximately Gaussian latent space linking encoders and generators, and relate it to real latents; as a proof of concept, we also explore a multi-view estimator that recovers a shared $k$-dimensional intersection subspace across models.

\item \textbf{Semantic structure in generative noise.}
We show that DDIM-inverted noise encodes rich semantics: linear probes on noise alone achieve strong attribute prediction across multiple diffusion models, closely matching foundation encoders.

\item \textbf{Controllable editing via linear directions.}
We enable faithful, interpretable edits by shifting along probe-derived directions in noise space, and show that a simple orthogonalization mitigates spurious entanglements, without architectural changes or fine-tuning.

\item \textbf{NoiseZoo dataset.}
We release \emph{NoiseZoo}: per-image DDIM-inverted noise paired with matched encoder embeddings for real images, enabling studies of generative-semantic correspondence.
\end{itemize}

\section{Related Work}
\textbf{Latent alignment and shared geometry.}
Despite architectural and objective differences, the latent spaces of VAEs~\cite{kingma2013auto}, GANs~\cite{goodfellow2014generative}, normalizing flows~\cite{rezende2015variational}, and diffusion models~\cite{ho2020denoising,song2020denoising} often exhibit surprising alignment. 
Empirically, several works show that simple linear mappings can translate between latent spaces~\cite{asperti2023comparing,badrinath2024all,lahner2023direct,maiorca2023latent,zhang2024emergence}, even across models trained independently or with different dimensionalities.
Other studies observe that cross-modal or cross-architecture representations remain compatible under shallow linear transforms~\cite{merullo2022linearly,bansal2021revisiting,kornblith2019similarity,liu2023visual,wang2024qwen2}.
A complementary direction seeks theoretical explanations for such alignment.
Conceptual frameworks like the Platonic Representation Hypothesis~\cite{huh2024platonic} and embedding translation~\cite{jha2025harnessing} argue that diverse models converge toward a shared latent description of the scene.
On the identifiability side, it was shown that InfoNCE can recover latent generative factors up to component-wise invertible transforms~\citep{zimmermann2021contrastive}, with follow-up work tightening this to linear identifiability and cross-encoder alignment~\cite{daunhawer2023identifiability,reizinger2024cross}.

However, these theoretical accounts assume a shared space without specifying its \emph{geometry}, while empirical alignment works reveal compatibility but offer no operational mechanism for \emph{using} the shared latent.
We instead propose that this shared space is not only present but approximately \emph{Gaussian}, making simple linear classification, semantic manipulation, and shared-space constructions natural operations that explicitly exploit its geometry.

\noindent \textbf{Gaussianity of representation spaces.}
Self-supervised learning implicitly encourages isotropy: contrastive learning spreads features uniformly on the hypersphere~\cite{wang2020understanding}, while redundancy-reduction methods decorrelate features~\cite{zbontar2021barlow,bardes2021vicreg}.
Whitening-based methods further produce Gaussianized embeddings~\cite{ermolov2021whitening}, and foundation model representations exhibit approximately Gaussian statistics~\cite{levi2025double, betser2025whitenedclip}.
Theory helps explain this trend: both contrastive and supervised training can recover latent factors up to linear transforms~\cite{daunhawer2023identifiability,reizinger2024cross,papyan2020prevalence}.
Additional work characterizes when representations exhibit Gaussian behavior~\citep{balestriero2025lejepa, balestriero2025gaussian,betser2026infonce}.
Prior work has shown that multi-modal representations exhibit a modality gap and often lie in lower-dimensional, anisotropic subspaces rather than being uniformly distributed~\citep{liang2022mind,shi2023towards,schrodi2024two,yaras2024explaining}. In this work, we focus on a single modality, namely the image modality.
These works, however, focus on encoder geometry only, whereas we place both encoders \emph{and} generative models under the same approximately Gaussian latent space.

\noindent \textbf{Semantic editing in generative latents.}
GANs enable editing along latent directions~\cite{shen2020interpreting,harkonen2020ganspace}, but diffusion models lack a persistent latent code.  
Recent approaches introduce editable subspaces~\cite{kwon2023diffusion,wang2025exploring}, or find directions via PCA, Jacobians or contrastive objectives~\cite{haas2024discovering,chen2024exploring,dalva2024noiseclr}.  
Null-text inversion~\cite{mokady2023null} and prompt-based manipulation~\cite{hertz2022prompt} improve controllability but do not expose explicit latent semantics. Recent work exploits approximate linearity of diffusion outputs for controllable sampling~\citep{song2025ccs}.
Unlike these methods, we operate directly in the \emph{noise space}, showing that it encodes semantic structure comparable to representation embeddings and enabling simple linear edits in noise space without prompt engineering or model fine-tuning.

\begin{table*}[htbp]
\centering
\renewcommand{\arraystretch}{1.05}
\setlength{\tabcolsep}{6pt}
\caption{Gaussianity measured via random 1D projections. For each model, we evaluate 5{,}000 projections of 250-sample subsets using Anderson-Darling (AD), D'Agostino-Pearson (DP), and Shapiro-Wilk (SW) tests (AD: lower is better $\downarrow$; DP, SW: higher is better $\uparrow$). AD\%, DP\% and SW\% denote the fraction of projections classified as Gaussian (AD $< 0.752$; DP and SW $p$-value $> 0.05$). Generative models approach the theoretical 95\% acceptance rate of Gaussian samples, encoders remain high, and non-Gaussian references perform substantially worse.}
\begin{adjustbox}{width=0.65\linewidth}
\begin{tabular}{l||cc|cc|cc}
\toprule
\multirow{2}{*}{\textbf{Model}} &
\multicolumn{2}{c|}{\textbf{Anderson--Darling (AD)}} &
\multicolumn{2}{c|}{\textbf{D'Agostino--Pearson (DP)}} &
\multicolumn{2}{c}{\textbf{Shapiro-Wilk (SW)}} \\
 &
\textbf{Avg.\ AD} $\downarrow$ & \textbf{AD \%} $\uparrow$ &
\textbf{Avg.\ DP} $\uparrow$ & \textbf{DP \%} $\uparrow$ &
\textbf{Avg.\ SW} $\uparrow$ & \textbf{SW \%} $\uparrow$ \\
\midrule

SD 1.5   & 0.387 & 96.00 & 0.504 & 95.16 & 0.491 & 95.02 \\
SD 2.1   & 0.381 & 95.80 & 0.496 & 94.76 & 0.497 & 94.96 \\
LCMv7    & 0.382 & 95.58 & 0.504 & 94.56 & 0.500 & 94.44  \\

\midrule

CLIP B16 & 0.451 & 89.50 & 0.430 & 88.88 & 0.424 & 88.22 \\
CLIP L14 & 0.429 & 91.90 & 0.441 & 90.62 & 0.442 & 90.16 \\
OpenCLIP B16   & 0.451 & 90.92 & 0.421 & 89.20 & 0.419 & 88.26 \\
OpenCLIP L14   & 0.429 & 92.20 & 0.443 & 90.80 & 0.440 & 90.04 \\
DINOv3   & 0.513 & 84.48 & 0.362 & 81.56 & 0.361 & 80.80 \\

\midrule

Delta & 171.73 & 0.00 & 0.001 & 0.12 & 0.089 & 8.58 \\
5D Uniform & 0.623 & 76.06 & 0.200 & 58.04 & 0.237 & 69.30 \\
Bimodal Gaussian & 16.380 & 15.88 & 0.074 & 16.36 & 0.068 & 15.16 \\

\bottomrule
\end{tabular}
\end{adjustbox}
\label{tab:latent_normality}
\end{table*}


\section{Universal Normal Embedding (UNE)}
Generative models and vision encoders share a key property: their latents exhibit approximately Gaussian structure.
Yet their capabilities differ, with encoders excelling at high-level semantic representations that support linear recognition and retrieval;
in contrast, generative models carry precise pixel-level information and can synthesize or reconstruct images.
For example, DDIM inversion can recover image-specific noise codes for a given diffusion model, but semantic editing in these models typically relies on external guidance (e.g., text prompts, architectural changes, or extra training) and remains limited without it.
Despite these differences in objective and usage, both families access the same data distribution (e.g., natural images) and, empirically, produce Gaussianized latent variables.
This complementarity motivates our central view: \emph{encoding and generation are two related directions over a shared latent Gaussian geometry}, which we formalize as the \emph{Universal Normal Embedding (UNE)} hypothesis.

\begin{tcolorbox}[myhypothesis, title=Hypothesis 1: Universal Normal Embedding (UNE)] 
Let $S$ be the data domain (e.g., natural images) with data sampled from a distribution $p$.
We posit that there exists a Gaussian latent space, the \emph{Universal Normal Embedding (UNE)}, which is $\mathcal{N}(0,I_D)$ of unknown dimension $D \in \mathbb{N}$ such that:

\begin{enumerate}
    \item There exists an invertible, information-preserving mapping between $S$ and the UNE.
    \item The UNE representation is \emph{simple} in the sense that there exists a set of semantic properties of the data, where each property is linearly separable.
\end{enumerate}
\label{hyp:une}
\end{tcolorbox}

\subsection{Induced Normal Embeddings}
In practice, models do not recover the full UNE for several reasons.
First, their latent dimensionalities differ, often chosen heuristically to balance performance and computational cost.
Second, variations in training objectives and architectures lead models to encode different aspects of the underlying information.
Third, the data modalities vary: for instance, CLIP is trained on paired image-text data, whereas DINO and most generative models are not.
Accordingly, both encoders and generative models realize an \emph{Induced Normal Embedding}: a model-specific latent space that is well-approximated by a noisy linear projection of the ideal UNE.
Some of the true latent structure is preserved, some dimensions may be discarded, and additional model-specific noise or redundancy may be injected.

Hence, all models are exposed to different parts of the ``true'' representation, varying due to different transforms and model-specific noise. An immediate consequence of our Hypotheses is that in the noiseless case ($\epsilon_i=0$ in \Cref{eq:ine}), if $C_i$ is invertible, any semantic property which is linearly separable in the UNE is also linearly separable in the INE.
Moreover, linear separability across multiple INEs suggests a shared low-dimensional space given by their intersection, preserving separability under linear projections.

\begin{tcolorbox}[myhypothesis, title=Hypothesis 2: Induced Normal Embeddings (INE)]
\label{hyp:ine}
Let $Z \sim \mathcal{N}(0, I_{D})$ be a random variable defined on the UNE.
We hypothesize that latent representations learned by modern encoders and generative models are induced by the UNE: for each model $i$ there exists a linear map $C_i \in \mathbb{R}^{d_i \times D}$ and a noise term $\epsilon_i$ such that its latent code $\hat{Z}_i$ satisfies
\begin{equation}
    \hat{Z}_i \;=\; C_i Z \;+\; \epsilon_i.
    \label{eq:ine}
\end{equation}
Consequently, each $\hat{Z}_i$ is (approximately) Gaussian and can be viewed as a noisy linear projection of $Z$.
\end{tcolorbox}

The UNE and INE hypotheses align with the Platonic Representation Hypothesis~\cite{huh2024platonic}, but extend it in several important ways. 
First, they explicitly state the Gaussianity of the underlying distribution, and state the correlation between the real distribution and the distribution of observations. 
Second, they unify not only encoders but both families of encoders and generative models. 
Lastly, since INEs are noisy linear projections of the UNE, and we have access to them, we can extrapolate properties such as linear separability.

\begin{figure*}[t]
    \centering
    \includegraphics[width=0.85\linewidth]{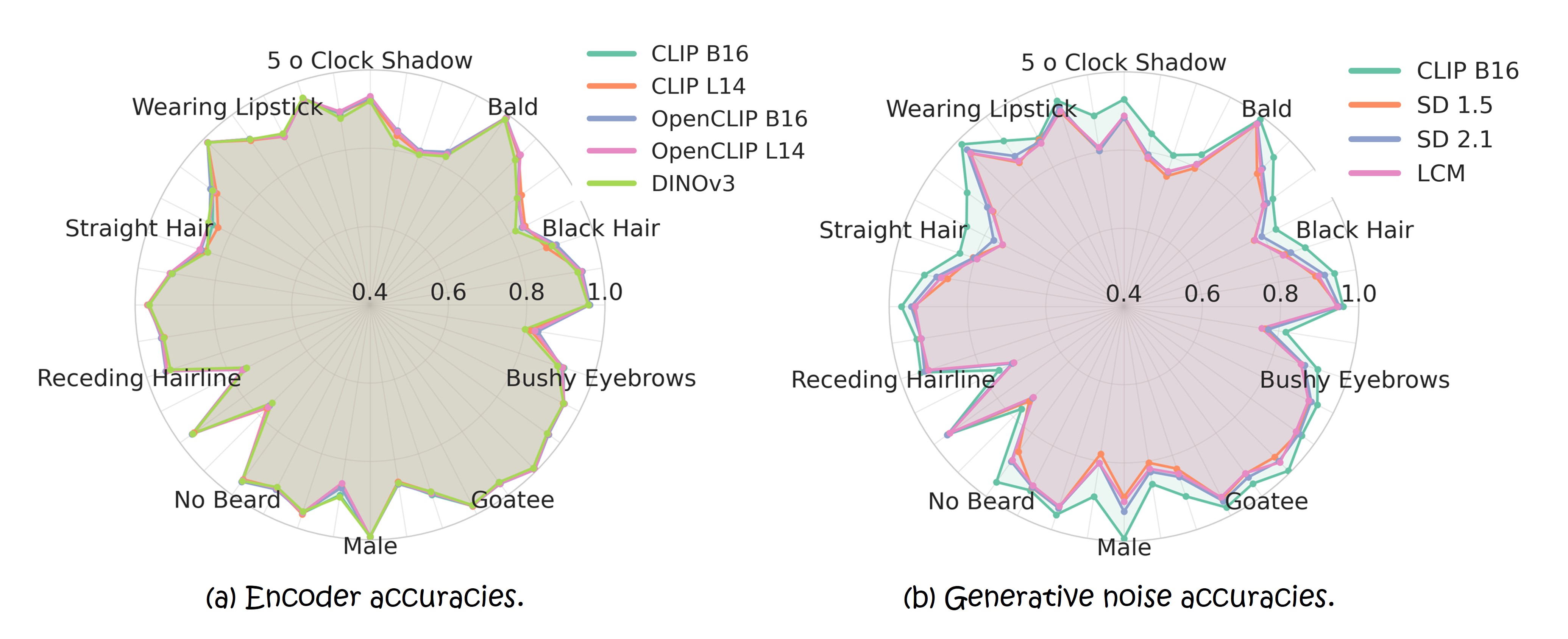}
    \caption{\textbf{Classification probing in latent spaces.}
    We train linear attribute classifiers (logistic regression) on latent representations from different models and evaluate accuracy on 40 CelebA attributes.
    (a) CLIP variants, OpenCLIP variants, and DINOv3 achieve nearly identical performance across attributes, demonstrating that semantic information is linearly accessible.
    (b) DDIM-inverted noise latents from SD~1.5, SD~2.1, and LCM achieve accuracy highly correlated with a strong encoder baseline (CLIP-B/16), despite originating from diffusion noise rather than semantic encoders.
    For clarity, only 10 representative attribute names are displayed.}
    \label{fig:spider_acc}
\end{figure*}

\noindent \textbf{Relation between INEs and UNE.}
INEs do not achieve the ideal Gaussian latent space. However, they contain a strong normal core: many latent directions behave as nearly Gaussian, while others capture redundancy or noise. While generative models (e.g., diffusion models) are trained to sample from a Gaussian latent prior, for representation models this happens without explicit normality constraints. Foundation encoders (CLIP, OpenCLIP, DINOv3~\cite{radford2021learning,mlf2021openclip, siméoni2025dinov3}) empirically push embeddings toward smooth and isotropic distributions. Consequently, both representation models and generative models naturally form latent spaces where ``Gaussian-like'' directions coexist with nuisance dimensions.
This phenomenon is experimentally verified in \Cref{tab:latent_normality}, where we assess eight models: three generators from the Stable Diffusion family~\cite{rombach2022high, luo2023latent} and five encoders (two CLIP variants, two OpenCLIP variants, and DINOv3). Across most models, more than 90\% of latent dimensions satisfy Gaussianity according to standard normality tests (Anderson-Darling and D'Agostino-Pearson~\citep{d1973tests, anderson1954test}), confirming that learned latents already approximate the normal structure predicted by the UNE hypothesis. Experimental details are provided in \Cref{sec:data}.

\subsection{Semantic directions}

A key property of Gaussian latent spaces is that Gaussian variables interact \emph{linearly}.
If a latent code $Z \in \mathbb{R}^d$ is standard normal and a semantic attribute
$Y \in \mathbb{R}$ is jointly Gaussian with $Z$, then the conditional expectation of $Y$
given the code is linear:
\begin{equation}
\label{eq:une_conditional_linear}
\mathbb{E}[\,Y \mid Z\,] = w^\top Z + b ,
\end{equation}
for some $w \in \mathbb{R}^d$ and $b \in \mathbb{R}$.
This follows directly from the closed form of the multivariate Gaussian conditional
distribution~\citep{bishop2006pattern}.
In this case, semantic variation corresponds to a linear direction in the latent space.

Many semantic attributes (e.g., age, height, smile intensity) behave approximately Gaussian when observed over a population: real-world measurements that arise from many small sources of variation tend to cluster around a mean and spread smoothly. In a Gaussianized latent space, such attributes align with linear directions $w$, making them effectively modeled by linear probes.
This motivates linear classifiers or regressors in latent space, which is well established for representation models (e.g., CLIP).
We further find that the same linear separability emerges in generative latent spaces such as DDIM-inverted noise, validated empirically in \Cref{fig:spider_acc}; details in \Cref{sec:class}.

\begin{figure*}[t]
    \centering
    \includegraphics[width=0.9\linewidth]{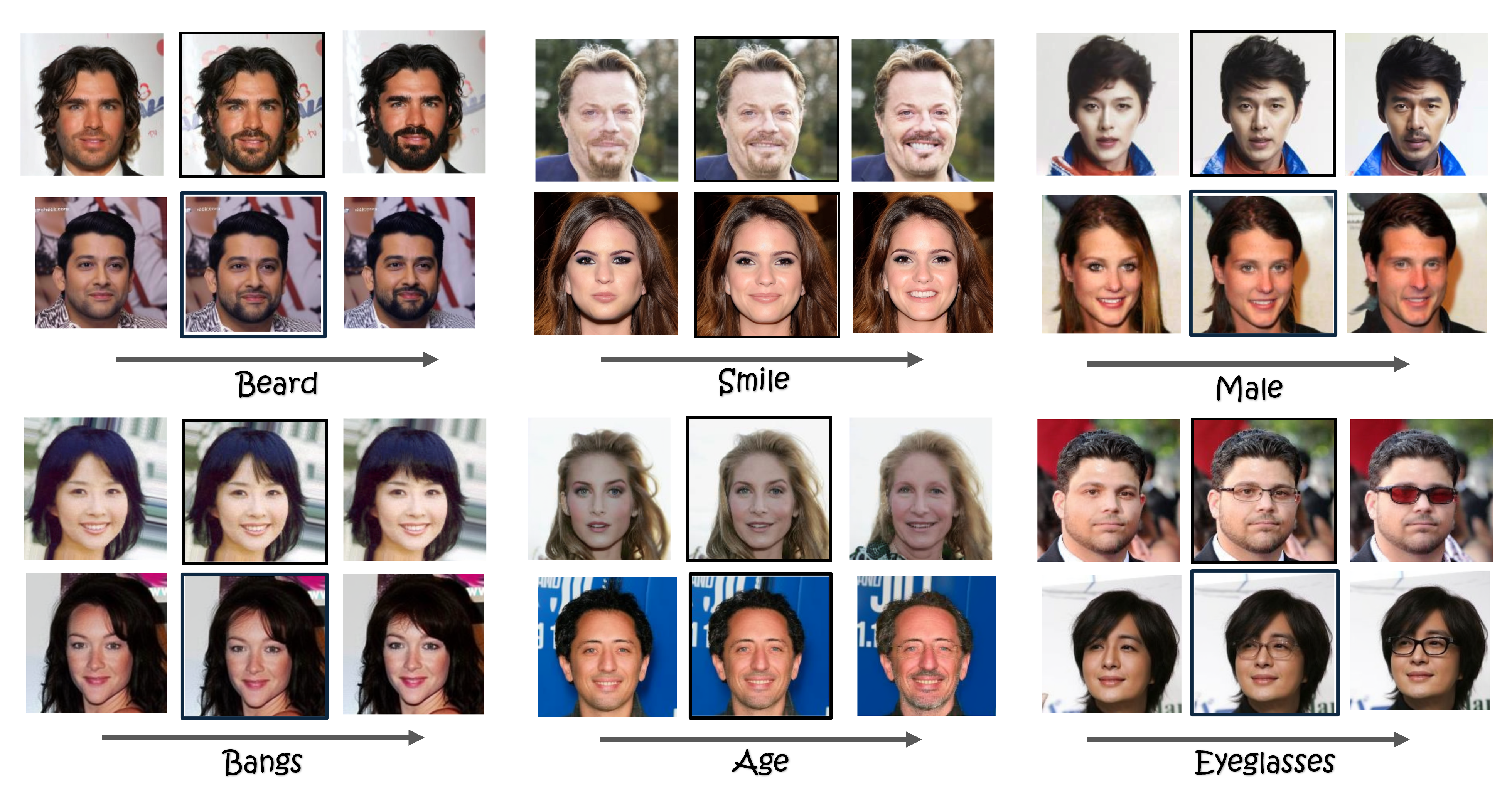}
    \caption{\textbf{Linear latent editing across semantic attributes.}
    For DDIM-inverted SD~1.5 latents, we move along linear classifier-derived semantic directions (\(\tilde{z} = z + \alpha w\)). 
    Each row shows decreasing (left) and increasing (right) attribute intensity as \(\alpha\) varies, middle image in each triplet is the original image.
    No prompts or model tuning are used; edits are controlled solely by linear shifts in the latent space.}
\label{fig:edit_example}
\end{figure*}

\noindent \textbf{Linear editing in latent spaces.}
With Gaussian latents and approximately Gaussian attributes, semantic changes often correspond to moving along linear directions.
This behavior is not limited to ideal UNEs: representation and generative models whose latents only approximate Gaussianity (e.g., diffusion noise through DDIM inversion) exhibit the same effect: linear probes reveal interpretable semantic directions.
In this setting, semantic editing corresponds to moving along a linear path: 

\begin{equation}
    \tilde{z} = z + \alpha\, w,
\label{eq:semantic_edit}
\end{equation}
where $w$ is the normal of the learned linear decision boundary and $\alpha$ controls edit strength. We demonstrate this simple linear editing in the DDIM-inverted space in \Cref{fig:edit_example}; details in \Cref{sec:edit}.

\noindent \textbf{Mitigating spurious features.}
Semantic directions are not always perfectly disentangled: a direction estimated for one
attribute may partially align with another, causing edits to change unintended properties.
To mitigate this, we edit along an \emph{orthogonalized} direction that removes the observed unintended changes by projecting the semantic direction into the null space of the spurious direction.
Formally, let $w_1,w_2\in\mathbb{R}^d$ be linear directions for two attributes. Changing attribute $B_1$ without affecting attribute $B_2$ can be formalized as:
\begin{equation}
\label{eq:orth_single}
\tilde w_1 \;=\; w_1 \;-\; \frac{w_2 w_2^{\top}}{w_2^{\top}w_2}\, w_1 .
\end{equation}
An illustration and examples of this simple mitigation strategy are presented in \Cref{fig:spurious_edit_example}; see details in \Cref{sec:edit}.

\subsection{Mapping between models and shared spaces}
\label{sec:map}
\noindent \textbf{Mapping between models.}
Each INE can be viewed conceptually as a noisy linear transformation of the same underlying normal latent space. Under this view, different models do not learn unrelated representations, they learn different linear embeddings of the same latent geometry. Therefore, moving from one model’s latent space to another should require only a linear mapping, with deviations attributable to noise or unused dimensions rather than fundamentally different structure.
While prior work has separately reported linear mapping within model families (encoders and generators), our hypotheses link both within a single latent framework.
This suggests a direct correspondence between generative latents
(e.g., DDIM-inverted diffusion noise) and representation embeddings (e.g., encoders).
We demonstrate this cross-family alignment in~\Cref{tab:mse-cos-accdrop}; experimental details are in~\Cref{sec:class}.

\noindent \textbf{Recovering the shared subspace of multiple INEs.}
Given $m$ models, each produces a learned latent representation of the same $n$ images.
Although these representations differ in dimensionality and contain noise or redundant directions, they are assumed to originate from the same underlying latent structure, the UNE (see \Cref{eq:ine}).
Our goal is to recover a shared $k$-dimensional latent space that all models ``agree on''.

Let $\hat{Z}_i \in \mathbb{R}^{n \times d_i}$ denote the latent codes of model $i$ (rows are samples, columns are centered features).
We seek a shared $k$-dimensional representation
$X \in \mathbb{R}^{n \times k}$
such that each model can linearly explain this same latent structure via some matrix
$A_i \in \mathbb{R}^{d_i \times k}$:
\begin{equation}
\label{eq:shared_space}
\hat{Z}_i A_i \;\approx\; X \qquad \forall\, i = 1,\ldots,m .
\end{equation}

\noindent Under the INE hypothesis (\Cref{eq:ine}), each $\hat{Z}_i$ is an approximately linear projection of the UNE.
We therefore treat $X$ as a $k$-dimensional proxy for this core space, and the $A_i$ as approximate ``inverse'' projections that recover $X$ from each INE.
This leads to the following objective:
\begin{equation}
\label{eq:main-une-optimization}
\begin{aligned}
\min_{X,\{A_i\}_{i=1}^m} \;&
\sum_{i=1}^{m} \|\, \hat{Z}_i A_i - X \,\|_F^2+\lambda_i\|A_i\|_F^2\\
\text{s.t.}\;& X^\top X = I, \quad 1^\top X = 0\;.
\end{aligned}
\end{equation}
The constraints $X^\top X = I$ and $1^\top X=0$ enforce centered features and identity covariance (up to scale) for the shared space, making $X$ an approximate instance of the form predicted by the UNE hypothesis. $\lambda_i > 0$ are chosen regularization parameters.
This objective corresponds to the MAXVAR formulation of Generalized Canonical Correlation Analysis (GCCA) \cite{kettenring1971canonical}.
It admits a closed-form solution: the matrix $X$ is obtained as the $k$ eigenvectors corresponding to the smallest eigenvalues of a matrix constructed from $\{\hat{Z}_i\}$ and $\{\lambda_i\}$. 
We note that the simplest form of GCCA sets $\lambda_i=0$ and drops the centering constraint $1^\top X=0$, but these nuances do not change the essential solution method. 
Our particular implementation is a hybrid approach that first optimizes $A_i$ in \Cref{eq:main-une-optimization} in closed form in terms of $X$, and then optimizes $X$ with $\lambda_i=0$ for all $i$.

Intuitively, this procedure identifies the \emph{intersection} of multiple INEs.
While it may not recover the full UNE, it extracts the portion of the latent structure consistently expressed across all models, and should be viewed as an initial construction among many possible alternatives.

\section{Experiments}
We curate \emph{NoiseZoo}, a dataset of per-image latents, and evaluate along three axes: 
(i) linear classification within and across latent spaces; 
(ii) controllable linear editing along probe-derived directions; 
(iii) recovery of a shared $k$-dimensional core via a multi-view estimator.

\begin{figure*}[t]
    \centering
    \includegraphics[width=0.95\linewidth]{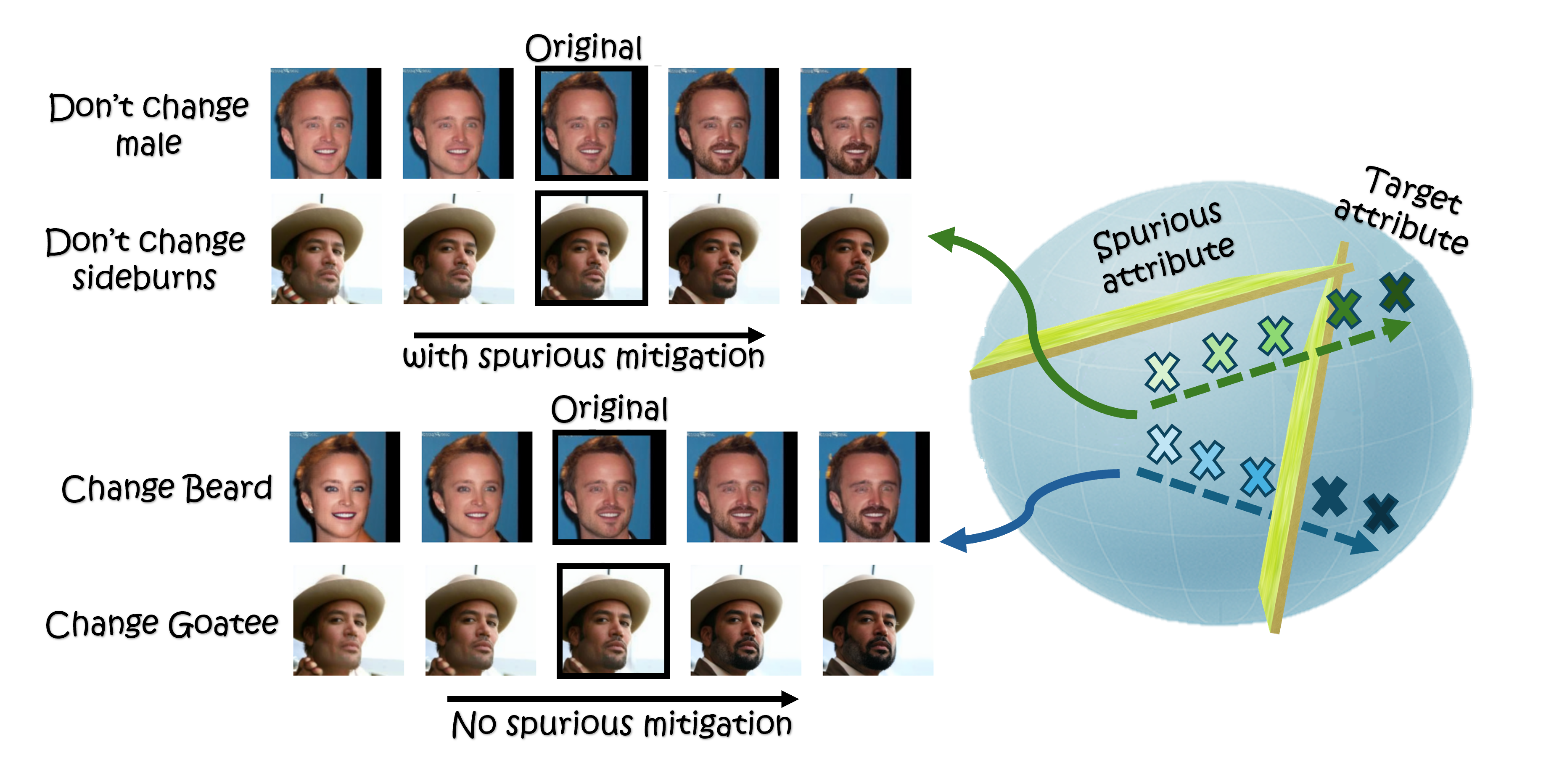}
    \caption{\textbf{Removing spurious attribute correlations.}
    Edits performed using the raw semantic direction (bottom) unintentionally modify a correlated attribute (e.g., adding a goatee also changes facial structure).
    Using the orthogonalized direction (top) from~\Cref{eq:orth_single} isolates the target attribute while suppressing the spurious one, yielding clean, disentangled edits.}    \label{fig:spurious_edit_example}
\end{figure*}

\begin{table*}[ht]
\centering
\caption{\textbf{Transferred latent evaluations.}
For each generative model, we linearly transfer its latents into encoder latent spaces and evaluate (i) geometric similarity (MSE, cosine similarity) and (ii) downstream attribute prediction accuracy using fixed encoder-trained classifiers.
An insignificant drop in accuracy (less than 0.3\%) and high similarity (high cosine similarity, low MSE) confirm that generative noise latents can be linearly aligned to encoder spaces while preserving predictive structure.}
\label{tab:mse-cos-accdrop}
\begin{adjustbox}{max width=0.75\textwidth}
\begin{tabular}{l|ccc|ccc|ccc}
\toprule
& \multicolumn{3}{c|}{\textbf{MSE} $\downarrow$}
& \multicolumn{3}{c|}{\textbf{Cosine Similarity} $\uparrow$}
& \multicolumn{3}{c}{\textbf{Accuracy Drop} (pp) $\downarrow$} \\
\cmidrule(lr){2-4}\cmidrule(lr){5-7}\cmidrule(lr){8-10}
\textbf{Model} & \textbf{CLIP B16} & \textbf{OpenCLIP L14} & \textbf{DINOv3}
               & \textbf{CLIP B16} & \textbf{OpenCLIP L14} & \textbf{DINOv3}
               & \textbf{CLIP B16} & \textbf{OpenCLIP L14} & \textbf{DINOv3} \\
\midrule
SD 1.5 & 0.072 & 0.215 & 0.143 & 0.8 & 0.73 & 0.55 & 0.20 & 0.23 & 0.23 \\
SD 2.1 & 0.073 & 0.219 & 0.146 & 0.80 & 0.72 & 0.54 & 0.14 & 0.19 & 0.20 \\
LCM    & 0.07 & 0.21 & 0.137 & 0.81 & 0.74 & 0.57 & 0.00 & 0.00 & 0.00 \\
\bottomrule
\end{tabular}
\end{adjustbox}
\end{table*}

\subsection{NoiseZoo construction}
\label{sec:data}
We use the CelebA~\citep{liu2015deep} validation set (\(\sim\) $19k$ images, split into $15k$ training and $4k$ test samples). 
For each image, we extract latent representations from five vision encoders:
CLIP ViT-L/14, CLIP ViT-B/16, OpenCLIP ViT-L/14, OpenCLIP ViT-B/16, and DINOv3~\citep{siméoni2025dinov3}. 
CLIP and OpenCLIP are contrastive image-text models trained on large-scale captioned datasets, whereas DINOv3 is trained purely on images using a self-supervised objective.
In addition, we obtain DDIM-inverted noise latents from three generative models in the Stable Diffusion family: SD~1.5, SD~2.1, and LCMv7~\citep{rombach2022high,luo2023latent}. 
SD~1.5 and SD~2.1 differ in training data and text encoders, while LCMv7 is trained under the Latent Consistency Model objective, which enables few-step sampling and induces a different geometry in the noise latent space.
Across models, encoder latents are moderately sized (500–1$k$ dimensions), whereas DDIM-inverted diffusion latents have much higher dimensionality (\(\sim\) 16$k$). Together, these models provide diverse generative and representation embeddings for the same underlying images.
This yields \emph{NoiseZoo}: a set of latents for every image. 
Details and examples are in Supp. Section~A.

\begin{figure*}[t]
    \centering
    \includegraphics[width=0.91\linewidth]{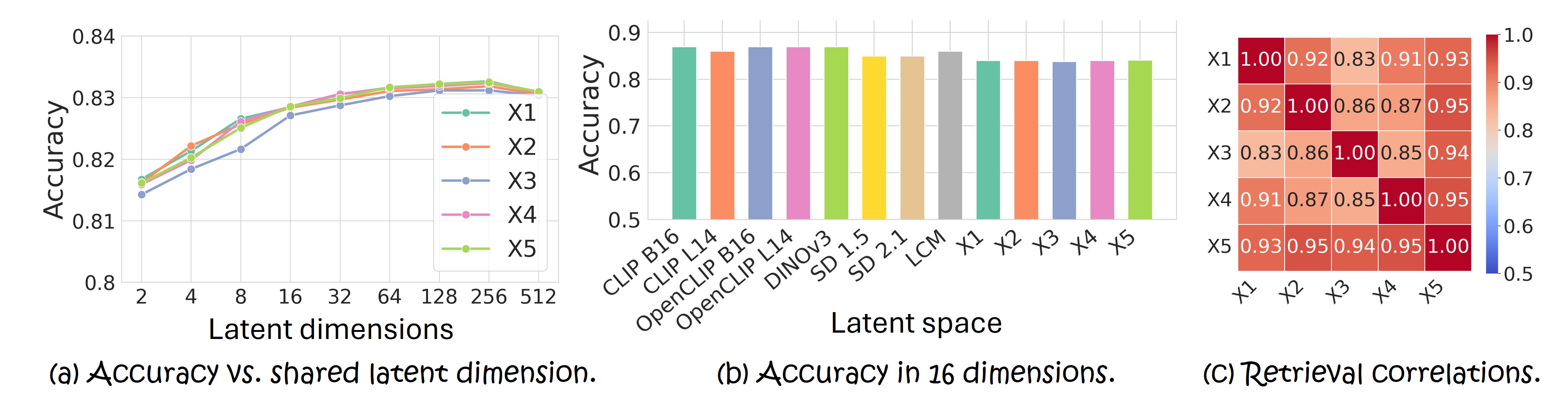}
    \caption{\textbf{Classification accuracy in shared latent spaces.}
    $X1$–$X4$ denote shared spaces computed from four latent sources, and $X5$ from six, with combinations detailed in \Cref{sec:shared_exp}.
    (a) Attribute classification accuracy as a function of latent dimension for shared spaces $X1$–$X5$, showing strong performance even at low dimensions.
    (b) Linear-probe accuracy at 16 dimensions in the PCA-reduced latent spaces of each model and in the low-dimensional shared spaces $X1$–$X5$, indicating that the shared space intersections retain comparable attribute information.
    (c) Retrieval-based analysis: pairwise correlations (Spearman rank correlation) of similarity vectors (computed over 10k latents) between the shared spaces $X1$–$X5$, demonstrating that they encode highly similar underlying structure.}
\label{fig:shared_acc}
\end{figure*}

\noindent \textbf{Assessing Gaussianity.}
We evaluate Gaussianity using Anderson-Darling, D'Agostino-Pearson, and Shapiro-Wilk tests on random 1D projections of the latent space~\citep{anderson1954test,shapiro1965analysis,d1973tests}. 
For each model, we sample 250 data points, compute 5{,}000 random projections, and report: 
(i) the average test statistic; and 
(ii) the fraction of projections that do \emph{not} reject normality.  
As shown in \Cref{tab:latent_normality}, generative models approach the theoretical 95\% acceptance rate of Gaussian samples and encoder representations score slightly lower but remain high. We additionally examine non-Gaussian reference distributions (delta distributions, low-dimensional uniform distributions, and bimodal Gaussians) as controls. These references perform substantially worse than both generative model and encoder representation spaces.

\subsection{Classification in latent spaces}
\label{sec:class}
For each model, we train logistic-regression classifiers (training details in Supp. Section A) for the 40 CelebA attributes using its training latents, and evaluate them on the corresponding test latents.
Attribute-wise accuracies for encoders and generative models are shown in \Cref{fig:spider_acc}, with CLIP ViT-B/16 overlaid on the generative panel for reference. 
Overall, DDIM-inverted noise latents yield attribute separability only slightly below that of leading encoders, with highly correlated per-attribute behavior across models.

\noindent \textbf{Cross-space transfer.}
To evaluate transferability, we learn ridge-regularized linear maps (Supp. Section A) from each generative latent space into three encoder spaces using the training split, and apply the \emph{fixed} encoder-trained classifiers to the mapped latents of the test set. 
Post-transfer performance is nearly unchanged, indicating that linear alignment suffices for downstream prediction. 
Mean square error (MSE), cosine similarity, and accuracy drops are reported in \Cref{tab:mse-cos-accdrop}, showing low error, high similarity, and minimal degradation.  
Additional results appear in Supp.\ Section~B.

\subsection{Linear editing}
\label{sec:edit}
Using the semantic directions from our linear classifiers, we edit DDIM-inverted latents and decode the modified samples. 
\Cref{fig:edit_example} shows edits across six CelebA attributes (SD~1.5), where varying intensity ($\alpha$ in \Cref{eq:semantic_edit}) smoothly increases or decreases attribute strength. 
Edits are local, controllable, and require no prompts or fine-tuning. 
To mitigate attribute entanglement, we use the orthogonalization in \Cref{eq:orth_single}; \Cref{fig:spurious_edit_example} demonstrates that it isolates the target attribute and suppresses spurious changes. 
We apply this procedure across all generative models and to CLIP ViT-L/14 (usable for synthesis guidance via UnCLIP~\citep{ramesh2022hierarchical}). 
Quantitative and additional qualitative results are provided in Supp.\ Section~B.

\subsection{Shared latent spaces}
\label{sec:shared_exp}
We compute shared latent spaces $Xi$ using the multi-view intersection method in \Cref{sec:map}. 
We consider four shared spaces constructed from four sources each and one from six sources; in every case, each shared space combines an equal number of encoder and generative model latents (see exact splits in Supp. Section~A).
\Cref{fig:shared_acc}a shows that all shared spaces achieve strong attribute classification in medium to high dimensions (32–512) and degrade in a similar manner when the dimension is reduced below 16.
We additionally apply PCA to each latent space individually; \Cref{fig:shared_acc}b shows that, at 16 dimensions, these PCA-reduced spaces reach similar classification performance to the shared spaces. 
Since each shared space is the intersection of its sources, it cannot contain more information than any single latent space.
This suggests that attribute information concentrates in a small set of shared directions.
To further test similarity between shared spaces, we perform a retrieval analysis on a subset of 10$k$ images: for each test latent, we measure its cosine similarity to all other latents and obtain a similarity vector. \Cref{fig:shared_acc}c reports Spearman rank correlations between these vectors across shared spaces, which are consistently high, indicating similar neighborhood structure.
Overall, these results provide a preliminary \emph{proof of concept} for the UNE hypothesis, suggesting that latent representations from both encoders and generative models retain highly similar underlying information.

\section{Conclusions}
We introduced the \emph{Universal Normal Embedding} hypothesis, proposing that generative and representation models approximate a shared Gaussian latent geometry where semantic factors correspond to linear directions. 
An immediate consequence of the UNE is that generators hold linearly separable semantics, in a similar manner to encoders.
Empirically, we demonstrated that DDIM-inverted noise codes and representation embeddings encode comparable semantic structure: attributes are linearly decodable in both, their probe predictions strongly agree, and noise-space classifier directions enable controllable edits without retraining or architectural changes. Current work primarily bridges the hypothesis with empirical findings. We plan to characterize the mechanisms that drive models toward UNE-like geometry in future studies. 
This work is a step toward a unification of representation learning and generative modeling, suggesting a shared geometric framework. 
We believe this viewpoint can guide both theoretical advances and the design of principled, interpretable generative systems.

\section*{Acknowledgments}
We would like to acknowledge  
support by the Israel Science Foundation (Grant 1472/23) and by the Ministry of Innovation, Science and Technology (Grant 8801/25).

{
    \small
    \bibliographystyle{ieeenat_fullname}
    \bibliography{combined}
}

\appendix
\newpage
\section*{Overview}
In this supplementary material document, we provide additional implementation and experimental details to ensure the full reproducibility (\Cref{supp:rep}). 
We also provide additional analyses and qualitative examples of our linear editing approach, along with experiments on an additional dataset (\Cref{supp: examples}).

\section{Reproducibility}
\label{supp:rep}
Code and the NoiseZoo dataset are available \href{https://rbetser.github.io/UNE/}{here}. Full implementation details are also available in the code repository.

\subsection{NoiseZoo construction details}
We constructed the NoiseZoo dataset by extracting latent representations for all 19,867 images in the CelebA~\citep{liu2015deep} validation split, without any filtering. The dataset includes latents from three Stable Diffusion variants (SD 1.5, SD 2.1, and LCM~\citep{rombach2022high, luo2023latent}), two CLIP variants (ViT-B/16 and ViT-L/14)~\citep{radford2021learning}, two OpenCLIP variants with the same architectures~\citep{mlf2021openclip}, and DINOv3 (ViT-L/16)~\citep{siméoni2025dinov3}. Additionally, the NoiseZoo dataset was randomly split into 15,893 training samples and 3,974 test samples. 

\par\vspace{4pt}\noindent \textbf{Stable Diffusion latents.}
Latent representations for diffusion models were obtained via DDIM inversion using the HuggingFace \textit{diffusers} library. All images were center-cropped and bilinearly resized to 512×512 prior to inversion. Inversion was performed with an empty text prompt, classifier-free guidance enabled, a guidance scale of 3.5 and a fixed random seed (42). SD 1.5 and SD 2.1 were inverted with 50 DDIM steps, while LCM used 150 steps and a DDIMScheduler (as the default LCM scheduler is DDPM-based and does not support inversion). For all Stable Diffusion models, we saved only the initial latent obtained from the inversion procedure. All Stable Diffusion latents have shape (4, 64, 64) and are flattened before all the experiments.

\par\vspace{4pt}\noindent \textbf{Encoder latents (CLIP, OpenCLIP, DINO).}
Encoder-based representations were obtained by passing each original CelebA image through the corresponding model using the model’s default preprocessing pipeline. For DINOv3 (ViT-L/16), images were center-cropped and resized to 224x224 before encoding. No additional normalization was applied. The embedding dimensions for each model are:
\begin{itemize}
    \item CLIP ViT-B/16: 512
    \item CLIP ViT-L/14: 768
    \item OpenCLIP ViT-B/16: 512
    \item OpenCLIP ViT-L/14: 768
    \item DINOv3 ViT-L/16: 768
\end{itemize}
Encoder embeddings were not normalized to unit norm.

\begin{figure}
    \centering
    \includegraphics[width=0.91\linewidth]{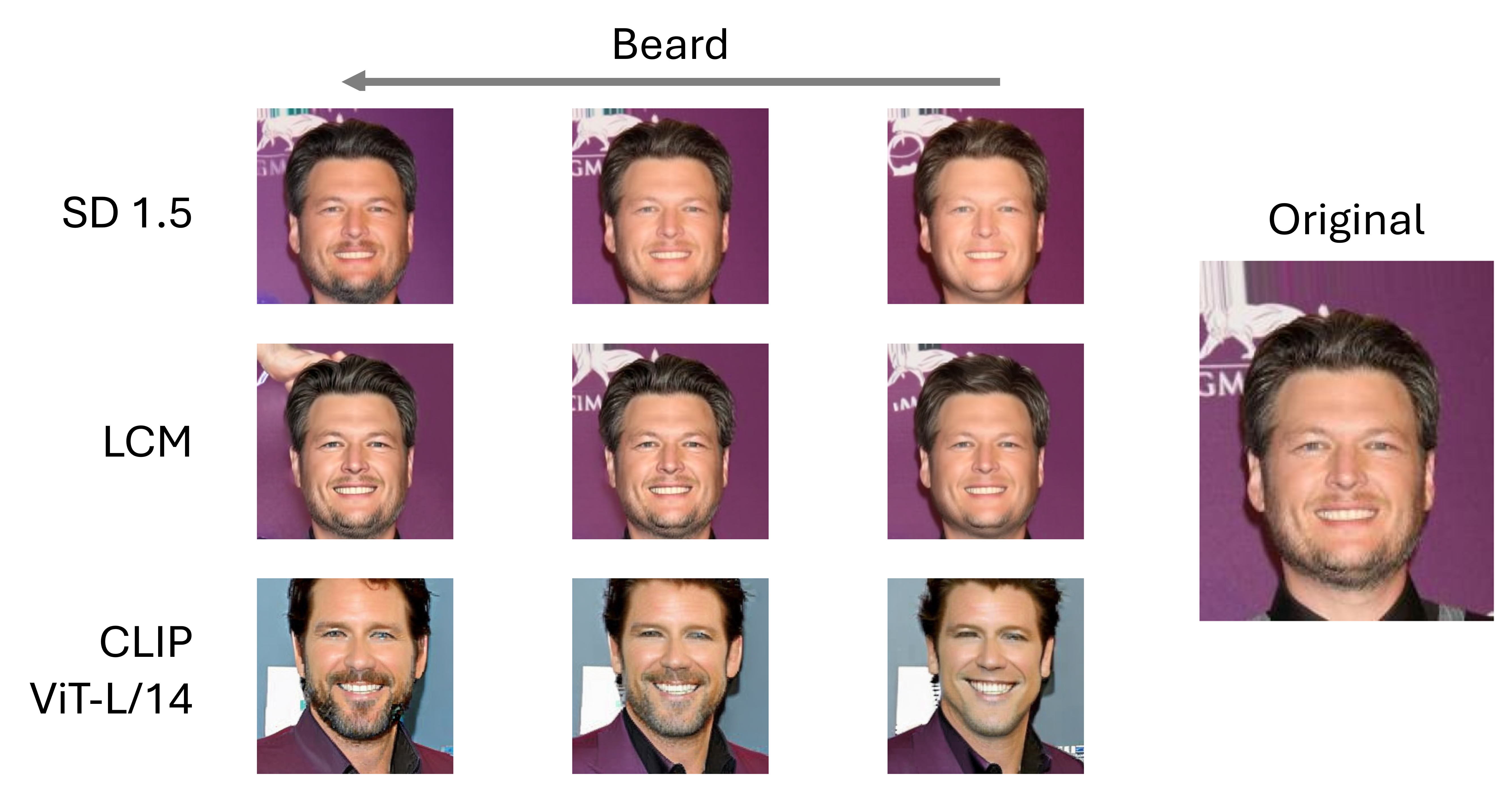}
    \caption{\textbf{Editing in different latent spaces.} The figure compares linear attribute editing across three latent spaces: two diffusion models and CLIP. The diffusion latents preserve the structure of the original image, so shifting along an attribute direction produces a modified version of the same image. In contrast, CLIP’s latent space is not invertible to the pixel domain, so reconstruction yields a newly synthesized image that matches the target attribute but does not reconstruct the input. This highlights the trade-off: CLIP offers strong semantic control but poor original image faithfulness.}
    \label{fig:editing_model_comp}
\end{figure}

\begin{figure*}
    \centering
    \includegraphics[width=0.91\linewidth]{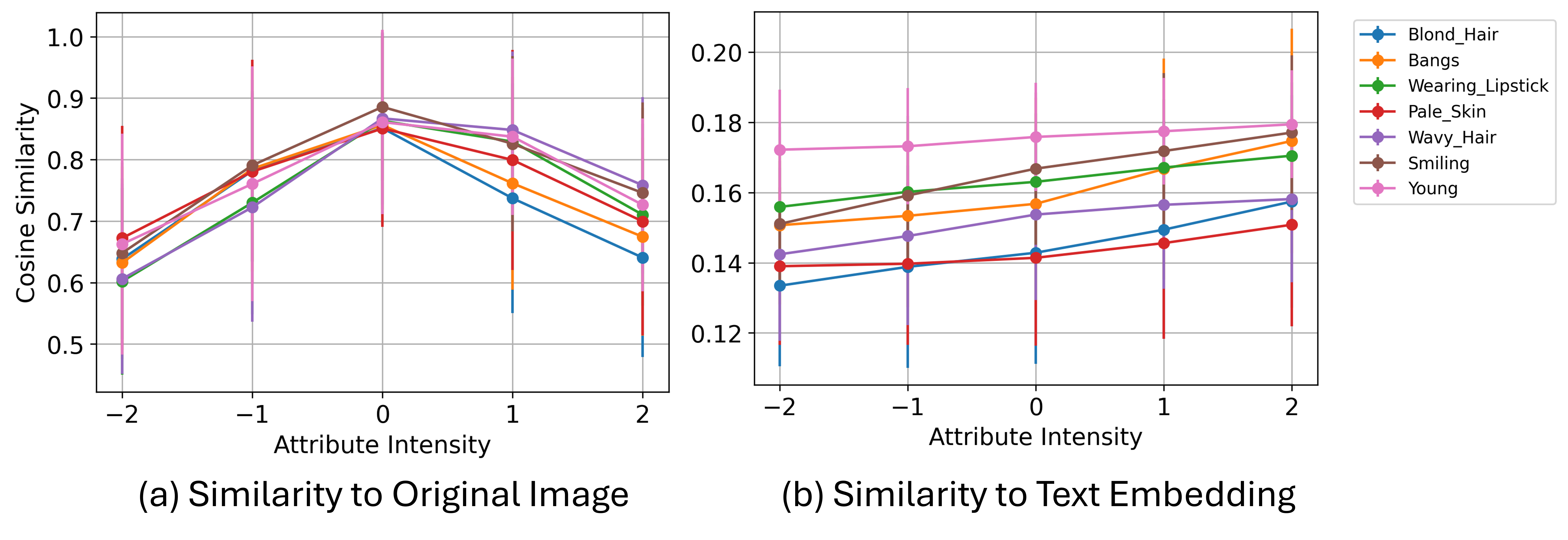}
    \caption{\textbf{Quantitative editing tests.} All measures are presented as a function of the edited \textit{attribute intensity}, a normalized measure derived from the distance between the resulting latent and the appropriate classifier's decision plane. Edits are performed on SD 1.5 latents. Note that an x-axis value of 0 does not indicate no editing, but corresponds to editing the latent to the classifier’s decision plane. Cosine similarities are measured between CLIP ViT-L/14 embeddings.
    (a) Cosine similarity between an edited image and the original image.
    (b) Cosine similarity between an edited image and CLIP text embeddings of the attribute's name.}
\label{fig:editing_quantitative}
\end{figure*}

\begin{figure*}[ht]
    \centering
    \includegraphics[width=\linewidth]{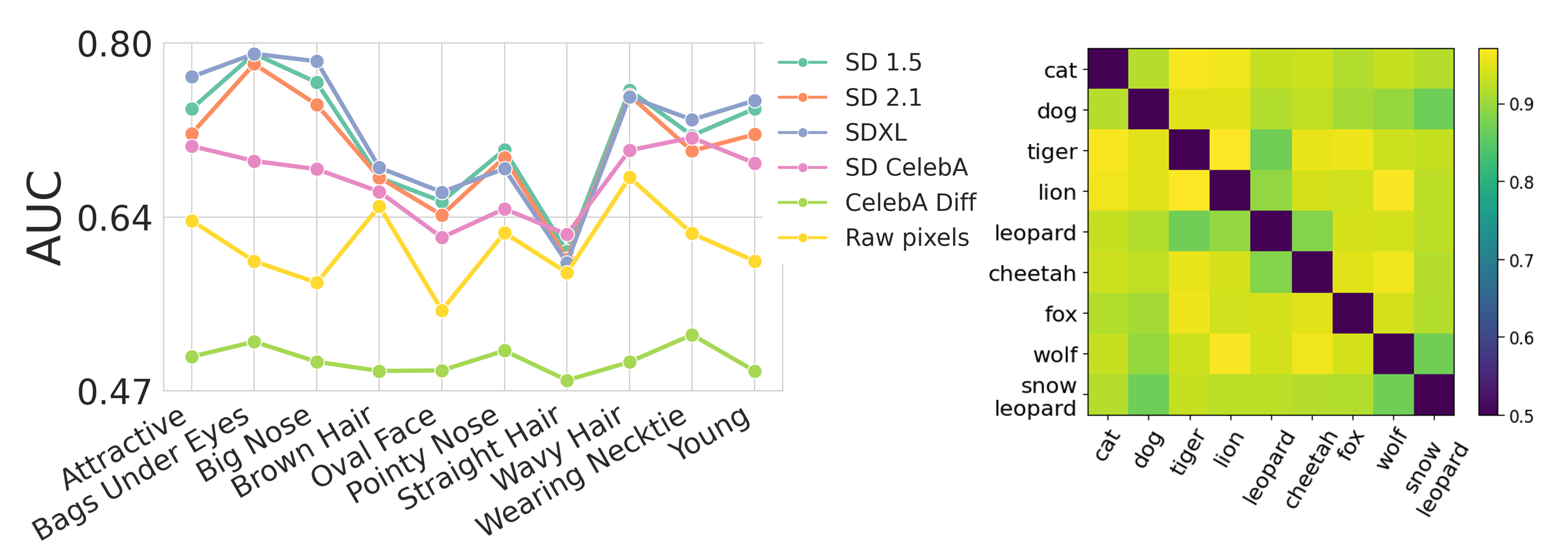}
    \caption{\textbf{Classification AUC.} Left: CelebA across different latent spaces. Right: AFHQ binary classification using SD 1.5 latents (AUC values).}
    \label{fig:auc}
\end{figure*}

\subsection{Experimental details}

\par\vspace{4pt}\noindent \textbf{Classification in latent space.}
For each feature set, the linear classifier consisted of a PCA projection, standard scaling, and an attribute-wise logistic regression stage. PCA was applied first (500 components for generative models and 310 for encoders), followed by standard scaling. Then, for each of the 40 attributes, a separate linear classifier was trained using scikit-learn’s LogisticRegression with the saga solver, L2 regularization, a maximum of 25 iterations, and 30 parallel jobs. Each attribute’s model forms one row of the overall weight matrix, with its corresponding bias term in the bias vector.

\par\vspace{4pt}\noindent \textbf{Cross-space transfer.}
We used ridge regression (scikit-learn's Ridge class) to learn a linear mapping between latent representations. The model was trained on paired samples in the training set, and evaluation (reported in Table 2 in the paper) was performed by applying a classifier trained in the target space to the translated test representations.

To ensure consistent regularization across different latent representations, the ridge penalty was scaled by the energy of the source features. The effective ridge penalty was set to $\alpha_{\text{eff}} = \alpha \frac{|| X_{\text{source}}||^2_F}{d}$, where $\alpha$ is the base regularization parameter, $X_{\text{source}}$ is the source feature matrix and $d$ is its dimensionality. We used $\alpha = 1.0$ in the reported results.

\par\vspace{4pt}\noindent \textbf{Shared latent spaces.}
The splits marked as $X1$-$X5$ in Figure 6 in the main paper are as follows:
\begin{itemize}[leftmargin=*]
    \item $X1$: SD~2.1, LCM, CLIP B/16, DINOv3
    \item $X2$: SD~1.5, LCM, OpenCLIP B/16, DINOv3
    \item $X3$: SD~1.5, SD~2.1, CLIP L/14, OpenCLIP B/16
    \item $X4$: SD~1.5, SD~2.1, CLIP L/14, DINOv3
    \item $X5$: SD~1.5, SD~2.1, LCM, CLIP L/14, OpenCLIP B/16, DINOv3
\end{itemize}

\begin{figure}[htpb]
    \centering
    \includegraphics[width=0.85\linewidth]{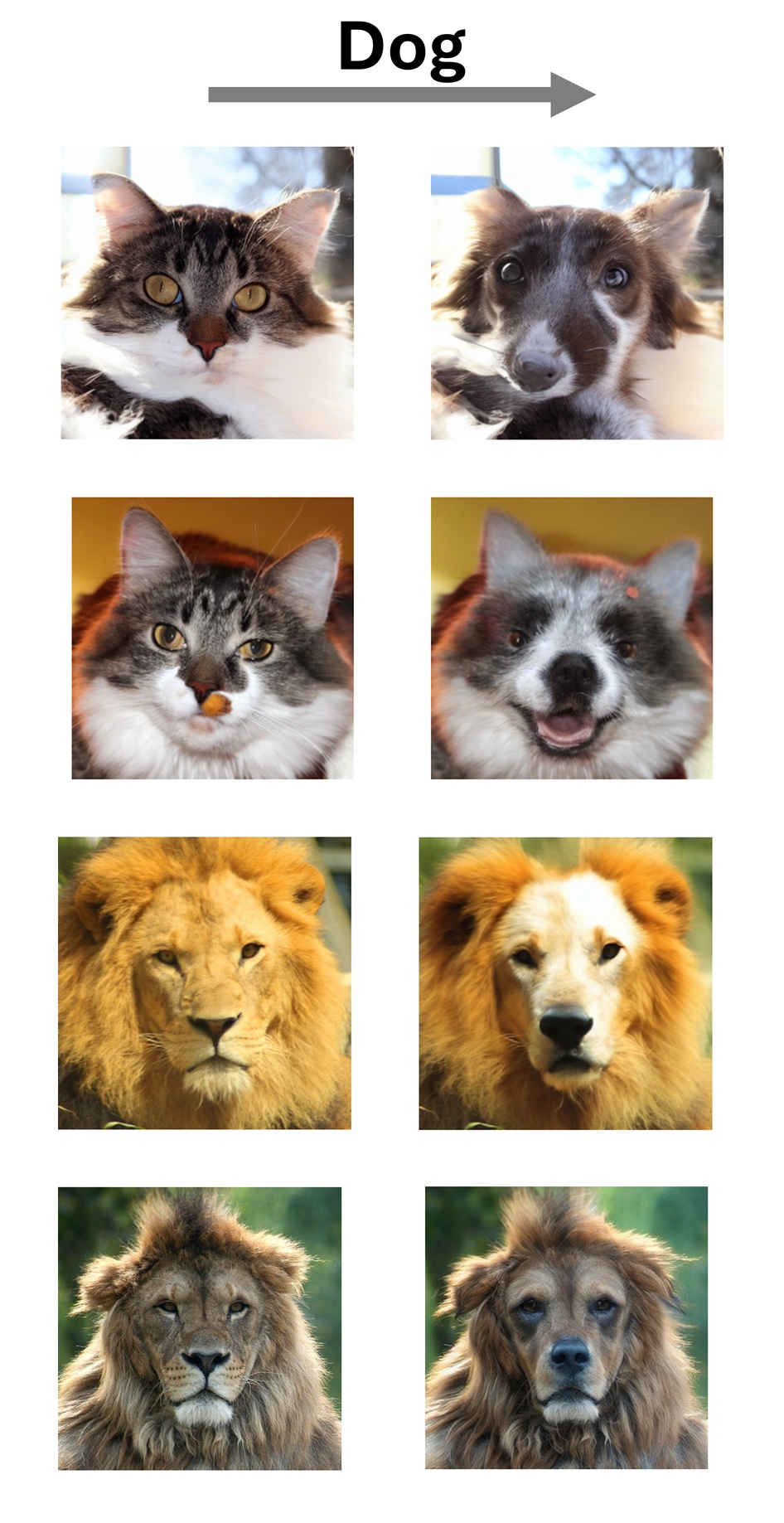}
    \caption{\textbf{Linear latent editing of animal faces.} We apply the method from Section 4.3 to the AFHQ dataset, which contains three categories: Cat, Dog, and Wild.}
    \label{fig:afhq_editing}
\end{figure}

\section{Editing Examples}
\label{supp: examples}

\subsection{Comparison of editing in different models} In \Cref{fig:editing_model_comp} we compare linear editing performed in the latent spaces of SD 1.5, LCM and CLIP ViT-L/14. As shown, diffusion latents allow faithful modification of the original image, whereas CLIP edits produce new images that satisfy the target attribute but do not preserve the input. The inversion of CLIP embeddings was done using the UnCLIP variant of Stable Diffusion~\cite{ramesh2022hierarchical,stabilityai2022sd21unclip}.

\subsection{Quantitative analysis of editing}
\Cref{fig:editing_quantitative} shows quantitative results of our editing method. The x-axis represents attribute intensity, a normalized measure derived from the distance to the classifier’s decision plane (x = 0 corresponds to editing to the decision plane, not no editing). Edits are performed on SD 1.5 latents. Panel (a) shows cosine similarity between the edited and original images, while panel (b) shows similarity between the edited image and the CLIP text embedding of the attribute name.

As intensity increases, similarity to the target attribute text embedding increases, indicating successful controlled editing. The similarity to the original image peaks at zero intensity. 

\subsection{Effect of model scale, conditioning, and pixel space}

In \Cref{fig:auc} (left), we compare linear attribute classification across different representations. As a baseline, we evaluate pixel space, as well as latent spaces from several diffusion models: Stable Diffusion 1.5 (SD 1.5), a version fine-tuned on CelebA (SD CelebA~\citep{realisticvisionv6}), a smaller unconditional model trained only on CelebA (CelebA Diff~\citep{celebahq}), and a larger model (SDXL~\citep{podell2023sdxl}).

Pixel-space representations yield substantially lower performance compared to generative latent spaces. The smaller model exhibits a clear degradation in linear separability, while increasing model scale (SDXL vs.\ SD 1.5/2.1) leads to only marginal improvements. Notably, fine-tuning on CelebA reduces linear separability even on the same dataset, highlighting the importance of broad and diverse training. Although CLIP influences training, it affects all samples uniformly at inference due to the use of empty-prompt DDIM inversion and generation.

\subsection{Evaluation on additional datasets}

To assess generalization beyond CelebA, we evaluate on AFHQ, which contains diverse animal faces~\citep{choi2020stargan}.  The collection of animal-face images spans three categories: Cat, Dog, and Wild. We add more granular labels to the Wild category using CLIP score with dominant class labels. As shown in \Cref{fig:auc} (right), semantic categories remain structured and linearly separable across species. We perform pairwise classification, with sub-categories defined via CLIP prompts.

To demonstrate that our latent editing procedure generalizes beyond human faces, we apply the method from Section 4.3 to this dataset.
We shift the latents of images along the classifier's direction, following the procedure described in the main paper. This produces realistic edits that preserve the structure of the original image, indicating that the learned directions capture shared high-level semantic information despite the dataset’s visual diversity.
Figure \ref{fig:afhq_editing} shows representative edits (towards the Dog class), illustrating attribute manipulation and confirming that the linearity assumption holds well in this domain.

\end{document}